\documentclass[runningheads]{llncs}
\usepackage{lmodern}
\usepackage[T1]{fontenc}
\usepackage{graphicx,verbatim}
\usepackage{amsmath,amssymb}
\usepackage{booktabs}
\usepackage[hidelinks]{hyperref}

\begin{document}
\let\oldref\ref
\renewcommand{\ref}[1]{{\NoHyper\oldref{#1}\endNoHyper}}
\title{Lymphocyte Mimicry Correction via Region-Level Tissue Reasoning and Unbalanced Optimal Transport}
\titlerunning{Lymphocyte Mimicry Correction via Loki-OT}

\author{Xiang Li\inst{1} \and
Yuqi Wang\inst{1} \and
Casey C. Heirman\inst{2} \and
Jihye Heo\inst{3} \and
Kyle J. Lafata\inst{1,2,4,5,6}}
\authorrunning{X. Li et al.}
\institute{Department of Electrical and Computer Engineering, Duke University, US \and
Medical Physics Graduate Program, Duke University, US \and
Department of Biomedical Engineering, Duke University, US \and
Department of Radiation Oncology, Duke University, US \and
Department of Radiology, Duke University, US \and
Department of Mathematics, Duke University, US}

\maketitle

\begin{abstract}
Cell mimicry arises when different cell types appear morphologically similar. Human pathologists resolve this ambiguity using surrounding tissue context, whereas current vision models either lack contextual reasoning (cell foundation models) or cannot operate at the cell level (pathology MLLMs). We present Loki-OT, which propagates region-level tissue reasoning to individual cell predictions via Unbalanced Optimal Transport, using MLLM-derived density priors as soft guidance for ambiguous cell reassignment. Loki-OT is motivated by the observation that pretrained cell foundation model features already encode discriminative information, including tissue context, but standard cell-level supervision fails to use tissue context effectively. The resulting transport plan is distilled into a lightweight student MLP classifier that learns context-aware decision boundaries within the pretrained feature space. On the independent TCGA-BRCA cohort, Loki-OT achieved lower patient-level MAE than the fully supervised in-domain PanopTILs classifier and improved F1 in epithelium-rich mimicry tissues, using 278 weak region-level MLLM estimates built on a general-domain cell foundation model. Code: \url{https://github.com/xiangli980/Lymphocyte_Mimicry_Correction_via_Loki_OT}.

\end{abstract}

\section{Introduction}
\label{sec:intro_related}

Tumor-infiltrating lymphocytes (TILs) are established prognostic and predictive biomarkers across multiple cancer types~\cite{ref_til_prognosis,ref_denkert,ref_resolution_gap}. However, accurate TIL quantification on H\&E-stained tissue remains challenging because morphologically similar cells can become difficult to distinguish in complex tissue environments, as recognized in pathology guidelines and pitfalls studies~\cite{ref_til_scoring,ref_stils_pitfalls}. We refer to this morphology-driven ambiguity as \emph{cell mimicry}, where non-lymphocyte nuclei visually resemble lymphocytes and contribute to systematic overestimation during automated cell classification. Recent analyses from the TIGER TIL challenge further identified confounding tissue morphology as an unresolved challenge for automated TIL quantification despite continued advances in model design and detection performance~\cite{ref_tiger_ai}.

Existing approaches address only part of this challenge. \textbf{Cell segmentation and classification models}, including CNN-based methods such as HoVerNet~\cite{ref_hovernet} and foundation-model-based methods such as CellViT~\cite{ref_cellvit}, achieve precise localization but rely primarily on local cellular morphology, which can fail when different cell types share similar morphology~\cite{ref_resolution_gap,ref_tiger_ai,ref_ocelot,ref_context_not_scale}. \textbf{Context-aware approaches}, including graph-based models~\cite{ref_cgcnet,ref_hactnet} and cell-on-tissue architectures such as OCELOT~\cite{ref_ocelot}, incorporate neighborhoods or tissue-structural features but still require dense cell-level annotations. In contrast, \textbf{pathology MLLMs} can reason about tissue composition---for example, recognizing that invasive tumor regions or normal glands should contain few lymphocytes---but operate only at the region level and cannot directly classify individual cells, whether generative~\cite{ref_pathchat}, contrastive-embedding-based~\cite{ref_conch}, or prompt-tuning-based~\cite{ref_top}. Consequently, a gap remains between region-level biological reasoning and cell-level prediction.

We formulate cell mimicry correction as learning cell-level predictions from region-level biological supervision. We present \textbf{Loki-OT} (LOgical Knowledge Injected OT), a framework that treats region-level tissue reasoning as a soft biological constraint and propagates it to individual cell predictions through Unbalanced Optimal Transport (UOT). Unlike multiple-instance learning (MIL)~\cite{ref_abmil} that learns bag-level representations through attention pooling, Loki-OT receives a region-level external target (the MLLM-derived lymphocyte density prior) and generates a transport plan for globally consistent cell-level assignments. Previous OT methods align empirical distributions, such as source and target domains~\cite{ref_deepjdot} or cellular populations~\cite{ref_pilot}. In this study, UOT aligns cell-level predictions with a biological prior describing the expected tissue composition. The resulting transport plan is distilled into a lightweight student classifier, requiring region-level reasoning only during training and using only the distilled classifier at inference (\textbf{Fig.~\ref{fig:pipeline}}).

\section{Method}
\label{sec:method}

\subsection{Cell Representation Encoding}

\begin{figure*}[t]
\centering
\includegraphics[width=\textwidth]{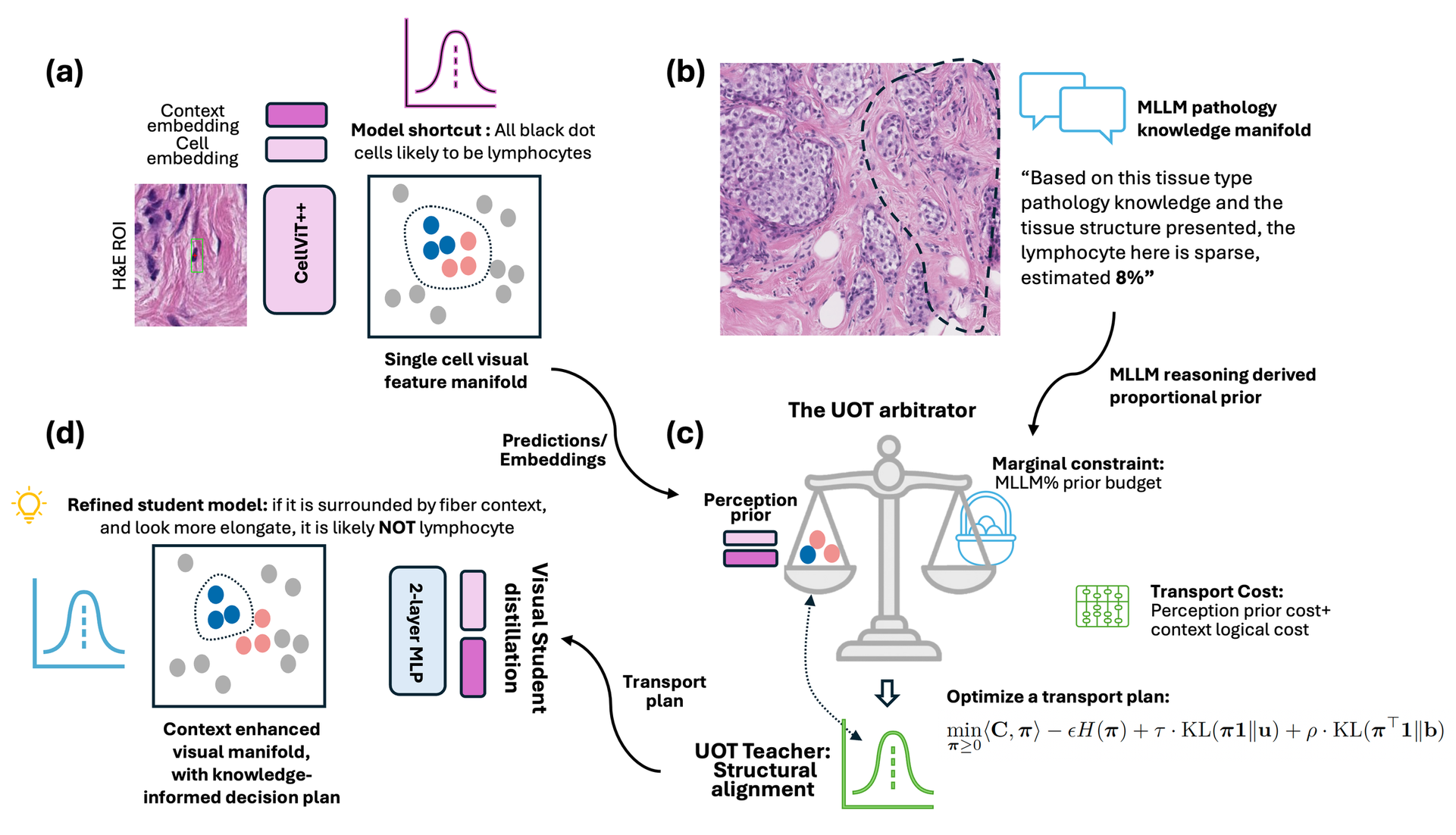}
\caption{Loki-OT pipeline. (a)~CellViT++ extracts per-cell morphological and context embeddings. (b)~An MLLM provides tissue-level lymphocyte density priors. (c)~The UOT arbiter balances perceptual evidence against biological priors to produce soft teacher labels. (d)~A lightweight MLP is distilled via a two-stage curriculum. At inference, only the distilled MLP runs, no MLLM or OT computation required.}
\label{fig:pipeline}
\end{figure*}

We employ CellViT++~\cite{ref_cellvit_plus_plus} with a SAM-H backbone to extract cell features (\textbf{Fig.~\ref{fig:pipeline}a}). For convenience, we use ``cell'' to denote each extracted nucleus-centered patch from CellViT++. For each cell $i$:
\begin{itemize}
\item \textbf{Morphological embedding} $\mathbf{z}_i \in \mathbb{R}^D$ ($D{=}1280$): the latent representation capturing nuclear morphology, from the last-layer ViT token of the encoder assigned to cell $i$.
\item \textbf{Context embedding} $\mathbf{c}_i = \frac{1}{|\mathcal{N}_i|} \sum_{j \in \mathcal{N}_i} \mathbf{z}_j \in \mathbb{R}^D$: mean-pooled ViT patch tokens $\mathbf{z}_j$ over $\mathcal{N}_i$, the expanded window of $\pm 5$ tokens ($\approx 80$ pixels) around cell $i$'s bounding box.
\item \textbf{Baseline prediction as perception prior} $\mathbf{a}_i \in \Delta^K$ ($K{=}2$): per-cell class probability from the model, collapsed to lymphocyte vs.\ non-lymphocyte.
\end{itemize}
Concatenating $[\mathbf{z}_i; \mathbf{c}_i]$ creates a context-enhanced feature space where morphologically identical cells in different tissue environments become separable.

\subsection{MLLM Knowledge Prior}

We query an MLLM with general pathology knowledge to estimate the expected lymphocyte density for each tissue region (\textbf{Fig.~\ref{fig:pipeline}b}). Given an ROI image $\mathbf{I}$, its tissue mask $\mathcal{M}$, and tissue subtype, the MLLM returns a region-level density prior $\mathbf{b} = \text{MLLM}(\mathbf{I}, \mathcal{M}, \text{tissue subtype}) \in \Delta^K$. The resulting prior serves as soft supervision for UOT during training only; at inference, the distilled student classifier operates without MLLM queries or tissue masks. We used a general-purpose instruction-following MLLM (Claude Sonnet 4.5) because it can better follow a structured density-estimation task in a masked region than a pathology-specific MLLM~\cite{ref_pathchat,ref_conch}. An example prompt is provided in App.~\ref{app:mllm_prompt}.
\subsection{Unbalanced Optimal Transport}

The cost matrix $\mathbf{C} \in \mathbb{R}^{N \times K}$ combines perceptual evidence with a logical penalty (\textbf{Fig.~\ref{fig:pipeline}c}): $C_{ik} = \text{Cost}_{\text{percept}}(i, k) + \text{Cost}_{\text{logic}}(i,k)$.
The perceptual cost uses the off-the-shelf baseline confidence, $\text{Cost}_{\text{percept}}(i,k) = 1 - a_{i,k}$, preserving high-confidence predictions. The logical cost encodes a meta-rule---\emph{lymphocytes in epithelium-rich regions are usually sparse}---implemented via cell-context similarity $s_i = \cos(\mathbf{z}_i, \mathbf{c}_i)$, and is added only in epithelium-rich tissue subtypes (T1, T3, T4):
\begin{equation}
\text{Cost}_{\text{logic}}(i,k) = \begin{cases} \text{ReLU}(s_i)^2 & k=\text{lymphocyte and cell } i \text{ lies in T1, T3, or T4} \\ 0 & \text{otherwise,} \end{cases}
\label{eq:logic}
\end{equation}
The logic cost assumes that a real lymphocyte should look visually distinct from its surrounding epithelium, so high cell--context similarity increases the cost of labeling a cell a lymphocyte. This is consistent with clinical TIL scoring guidelines that treat intratumoral lymphocytes as rare relative to stromal lymphocytes~\cite{ref_stils_pitfalls}, motivating the restriction of this penalty to epithelium-rich tissue subtypes (T1, T3, T4).

We formulate cell re-assignment as a UOT problem with uniform source mass $\mathbf{u}=\mathbf{1}_N/N \in \Delta^N$ and target class marginal $\mathbf{b} \in \Delta^K$:
\begin{equation}
\min_{\boldsymbol{\pi} \geq 0} \langle \mathbf{C}, \boldsymbol{\pi} \rangle - \epsilon H(\boldsymbol{\pi}) + \tau \cdot \text{KL}(\boldsymbol{\pi}\mathbf{1} \| \mathbf{u}) + \rho \cdot \text{KL}(\boldsymbol{\pi}^\top\mathbf{1} \| \mathbf{b})
\label{eq:uot}
\end{equation}
where $H(\boldsymbol{\pi})$ is the entropy regularizer that enables an efficient Sinkhorn-Knopp solution~\cite{ref_sinkhorn}. Unlike balanced OT, UOT allows deviation from both distributions through KL penalties ($\tau{=}\rho{=}3.0$), providing a mathematical buffer for MLLM uncertainty. We set $\epsilon=0.1$. The perception prior enters through $\mathbf{C}$ (via $\text{Cost}_{\text{percept}}$), while the MLLM density $\mathbf{b}$ is the target marginal; the transport plan $\boldsymbol{\pi}$ balances the two, preserving confident per-cell predictions except where the region must match the MLLM density.

\subsection{Two-Stage Training Strategy}
\label{sec:method_training}

Direct UOT distillation is unstable (\textbf{Fig.~\ref{fig:pipeline}d}) because transport assignments are sparse and differ substantially from baseline predictions. We therefore adopt a two-stage curriculum~\cite{ref_curriculum}: Stage~1 operates at region level, solving all $N$ cells within a tissue region jointly with $\mathbf{b}$ as the marginal constraint, while Stage~2 refines at cell level, distilling per-cell UOT assignments $\boldsymbol{\pi}_i$ as individual targets. UOT correction is applied only when the baseline lymphocyte density exceeds the MLLM estimate by at least $10\%$, i.e., $\Delta = \bar{a}_{\text{lymph}} - b_{\text{lymph}} \geq 10\%$; otherwise, predictions are preserved through self-distillation. The $10\%$ threshold is set based on the MLLM prior's density error (${\approx}\,11.7\%$).

\noindent\textbf{Stage~1 (region-proportional supervision, Context-soft version)} serves as warm-up initialization, using MLLM density as region-level (label-proportion) supervision~\cite{ref_llp}. The region-averaged prediction $\bar{f}_\theta = \frac{1}{|R|}\sum_{i\in R} f_{\theta,\text{lymph}}(\mathbf{x}_i)$ is regularized toward the MLLM density $b_{\text{lymph}}$ where $\Delta \geq 10\%$, and otherwise preserves baseline predictions:
\begin{equation}
\mathcal{L}_1 = \begin{cases}
\mathrm{ReLU}(\bar{f}_\theta - b_{\text{lymph}} - 0.05)^2 & \Delta \geq 10\% \\
\text{KL}(\mathbf{a}_i \| f_\theta) & \Delta < 10\%
\end{cases}
\end{equation}
The one-sided hinge with a $0.05$ dead-zone suppresses only over-prediction and tolerates small density errors.

\noindent\textbf{Stage~2 (UOT teacher distillation, final Loki-OT version)} distills cell-level teacher targets $\boldsymbol{t}_i$ through a temperature-softened KL term ($T{=}4$) combined with a hard-label cross-entropy anchor on the teacher's argmax to prevent probability collapse~\cite{ref_distill_survey,ref_hinton_distill}:
\begin{equation}
\mathcal{L}_2 = \tfrac{1}{2}\,\text{KL}(\boldsymbol{t}_i \| f_\theta) + \tfrac{1}{2}\,\text{CE}(f_\theta,\, \arg\max_k t_{i,k}), \quad
\boldsymbol{t}_i = \begin{cases} \boldsymbol{\pi}_i & \Delta \geq 10\% \\ f_{\theta_1} & \Delta < 10\% \end{cases}
\end{equation}
The teacher target $\boldsymbol{t}_i$ is the UOT transport assignment $\boldsymbol{\pi}_i$ in corrected regions ($\Delta \geq 10\%$) and falls back to the Stage-1 prediction $f_{\theta_1}$ elsewhere, so distillation is applied selectively.

\section{Experiments}
\label{sec:experiments}

\subsection{Experimental Setup}

\noindent\textbf{Datasets.} We use the TIGER challenge dataset~\cite{ref_tiger_ai} with expert nuclear annotations and tissue labels. Train: 135 ROIs ($1024{\times}1024$ pixels, ${\sim}$80K nuclei, 278 (ROI, subtype) regions) from 2 hospitals. Test: TIGER's TCGA-BRCA subset (1,741 ROIs, median $143{\times}143$ pixels, 64K nuclei, 124 patients) with the same annotation format and tissue masks, but zero hospital overlap. Both splits share 7 tissue subtypes (T1--T7) but differ in composition: training is T1-dominated ($43.5\%$ of cells) with minimal T6 ($3.3\%$), whereas the test set is T6-heavy ($32.8\%$). Within-tissue lymphocyte density is broadly comparable across splits (\textbf{Table~\ref{tab:density}}), so this is a composition shift rather than a density shift. MLLM (Claude Sonnet 4.5) provides weak supervision at training only; against training GT, its density estimates achieve Pearson $r{=}0.58$, an MAE of $11.7\%$, and $+6.6\%$ bias, averaged across all tissue subtypes (per-tissue MLLM\% in \textbf{Table~\ref{tab:density}}). The statistics show MLLM is coarse but directionally closer to GT than Lizard. We hypothesize that this relative direction is the key signal for the error correction.

\begin{table}[t]
\caption{Lymphocyte density (\%) by tissue subtype. Nuclei columns show train/test counts. Lizard over-predicts in epithelium-rich regions (T1, T3, T4); MLLM is coarse but directionally closer to GT than Lizard.}
\label{tab:density}
\centering
\setlength{\tabcolsep}{3pt}
\begin{tabular}{@{}l|rrrr|rrr@{}}
\toprule
 & \multicolumn{4}{c|}{Train (2 hospitals)} & \multicolumn{3}{c}{Test (TCGA-BRCA)} \\
Tissue subtype & Nuclei & GT\% & MLLM\% & Lizard\% & Nuclei & GT\% & Lizard\% \\
\midrule
T1 Invasive tumor & 35K & 1.3 & 13.9 & 28.6 & 26K & 2.4 & 24.4 \\
T2 Tumor-assoc.\ stroma & 18K & 20.3 & 20.8 & 31.0 & 14K & 23.2 & 38.3 \\
T3 In-situ tumor & 7K & 0.2 & 11.0 & 35.8 & 2K & 0.9 & 14.3 \\
T4 Healthy glands & 7K & 1.8 & 2.8 & 31.2 & 0.2K & 7.0 & 36.0 \\
T5 Necrosis & 0.6K & 5.4 & 0.3 & 38.1 & 0.3K & 1.5 & 52.1 \\
T6 Inflamed stroma & 3K & 52.6 & 64.4 & 41.1 & 21K & 61.6 & 67.8 \\
T7 Rest & 10K & 9.3 & 15.5 & 27.3 & 0.3K & 9.4 & 19.8 \\
\bottomrule
\end{tabular}
\end{table}

\noindent\textbf{Baselines and Ablations.} All methods share the same frozen CellViT++~\cite{ref_cellvit_plus_plus} (SAM-H) backbone and lightweight MLP architecture --- the only variable is the supervision signal. External baselines use cell-level GT labels: (1)~\textbf{Lizard}: MLP trained on large-scale general-domain colorectal (colon) cell labels, applied zero-shot to breast; (2)~\textbf{PanopTILs}: MLP retrained on 800K in-domain TCGA-BRCA nuclei with cell-level annotation --- in-domain, but its annotations over-represent epithelium-rich regions. (3)~\textbf{Context-soft}: Stage~1 training with 278 region-level density estimates as a proportional regularizer, serving both as the Stage-1 step ablation and as the standalone region-proportional-supervision comparison method; (4)~\textbf{OT/UOT-Teacher}: balanced/unbalanced OT solver outputs (teacher only, not deployable); (5)~\textbf{Loki-OT}: distilled student from two-stage curriculum, the final deployable model. Ablation is conducted on the training set, as OT/UOT-Teacher are solver outputs within the training pipeline.

\noindent\textbf{Implementation.} CellViT++ with frozen SAM-H encoder. Stage~1: 20 epochs, lr$=10^{-4}$; Stage~2: 5 epochs, lr$=10^{-5}$; Adam. UOT and distillation hyperparameters were selected on the training set against the MLLM density gap. Cell-level ground truth was not used in model training.

\noindent\textbf{Metrics.} Precision, recall, and F1 for lymphocyte classification are micro-averaged globally over all test-set cells; cell-level decisions (TP, FP, FN) use argmax, while AUC and AP use the raw softmax lymphocyte probability. Patient-level MAE first accumulates predicted and ground-truth lymphocyte counts across each patient's ROIs, then computes the per-patient absolute count difference, and finally averages this error over the 124 test patients. We report MAE in Table~\ref{tab:main} at patient level because TIL scoring is a per-patient biomarker~\cite{ref_til_scoring}---the ROI is the annotation unit but the patient is the clinical decision unit. Per-subtype MAE (App.~\ref{app:per_tissue_mae}) is computed the same way but aggregated at the ROI level, since most patients contribute only a few ROIs of any given subtype. For ablation, selectivity $=$ FP$\downarrow$\%\,/\,TP$\downarrow$\% measures how selectively false positives are removed while true positives are preserved.

\subsection{Main Results}
\label{sec:results}

\begin{table}[t]
\caption{TCGA-BRCA independent cohort (124 patients). F1, recall, and precision are globally micro-averaged; MAE is computed at the patient level. Point estimate [95\% bootstrap CI, $N{=}1000$]. Bold: best per column. Wilcoxon with Holm correction: Loki-OT vs.\ PanopTILs MAE $p{=}0.031$.}
\label{tab:main}
\centering
\setlength{\tabcolsep}{3pt}
\renewcommand{\arraystretch}{1.1}
\footnotesize
\begin{tabular}{@{}lcccccc@{}}
\toprule
Method & F1$\uparrow$ & Recall$\uparrow$ & Prec$\uparrow$ & AUC$\uparrow$ & AP$\uparrow$ & MAE$\downarrow$ \\
\midrule
Lizard     & \shortstack{0.526\\{\scriptsize[.492,.557]}} & \shortstack{0.743\\{\scriptsize[.707,.776]}} & \shortstack{0.393\\{\scriptsize[.354,.430]}} & \shortstack{0.787\\{\scriptsize[.768,.805]}} & \shortstack{0.544\\{\scriptsize[.509,.578]}} & \shortstack{83.8\\{\scriptsize[65.6,104.7]}} \\[4pt]
Context-soft & \shortstack{\textbf{0.579}\\{\scriptsize[.547,.606]}} & \shortstack{\textbf{0.760}\\{\scriptsize[.721,.797]}} & \shortstack{0.436\\{\scriptsize[.396,.473]}} & \shortstack{\textbf{0.823}\\{\scriptsize[.806,.838]}} & \shortstack{0.568\\{\scriptsize[.537,.603]}} & \shortstack{68.5\\{\scriptsize[54.4,84.4]}} \\[4pt]
Loki-OT    & \shortstack{0.543\\{\scriptsize[.512,.573]}} & \shortstack{0.637\\{\scriptsize[.599,.675]}} & \shortstack{0.455\\{\scriptsize[.418,.496]}} & \shortstack{0.809\\{\scriptsize[.792,.826]}} & \shortstack{0.554\\{\scriptsize[.521,.587]}} & \shortstack{\textbf{46.0}\\{\scriptsize[35.5,56.9]}} \\[4pt]
PanopTILs  & \shortstack{0.474\\{\scriptsize[.435,.511]}} & \shortstack{0.430\\{\scriptsize[.380,.477]}} & \shortstack{\textbf{0.561}\\{\scriptsize[.516,.604]}} & \shortstack{0.816\\{\scriptsize[.795,.835]}} & \shortstack{\textbf{0.578}\\{\scriptsize[.544,.612]}} & \shortstack{58.3\\{\scriptsize[45.4,71.5]}} \\
\bottomrule
\end{tabular}
\end{table}

\noindent\textbf{Does MLLM context supervision improve over Lizard?}
Context-soft improves every per-patient metric over Lizard (\textbf{Table~\ref{tab:main}}), with the
largest gain in MAE.
The results show that a coarse, imperfect MLLM prior (which encodes tissue-level reasoning) could reduce systematic
over-counting by directing Stage~1 to suppress density in regions where Lizard's false positive predictions
far exceed the tissue-level estimate.

\begin{figure}[!t]
\centering
\includegraphics[width=0.95\textwidth]{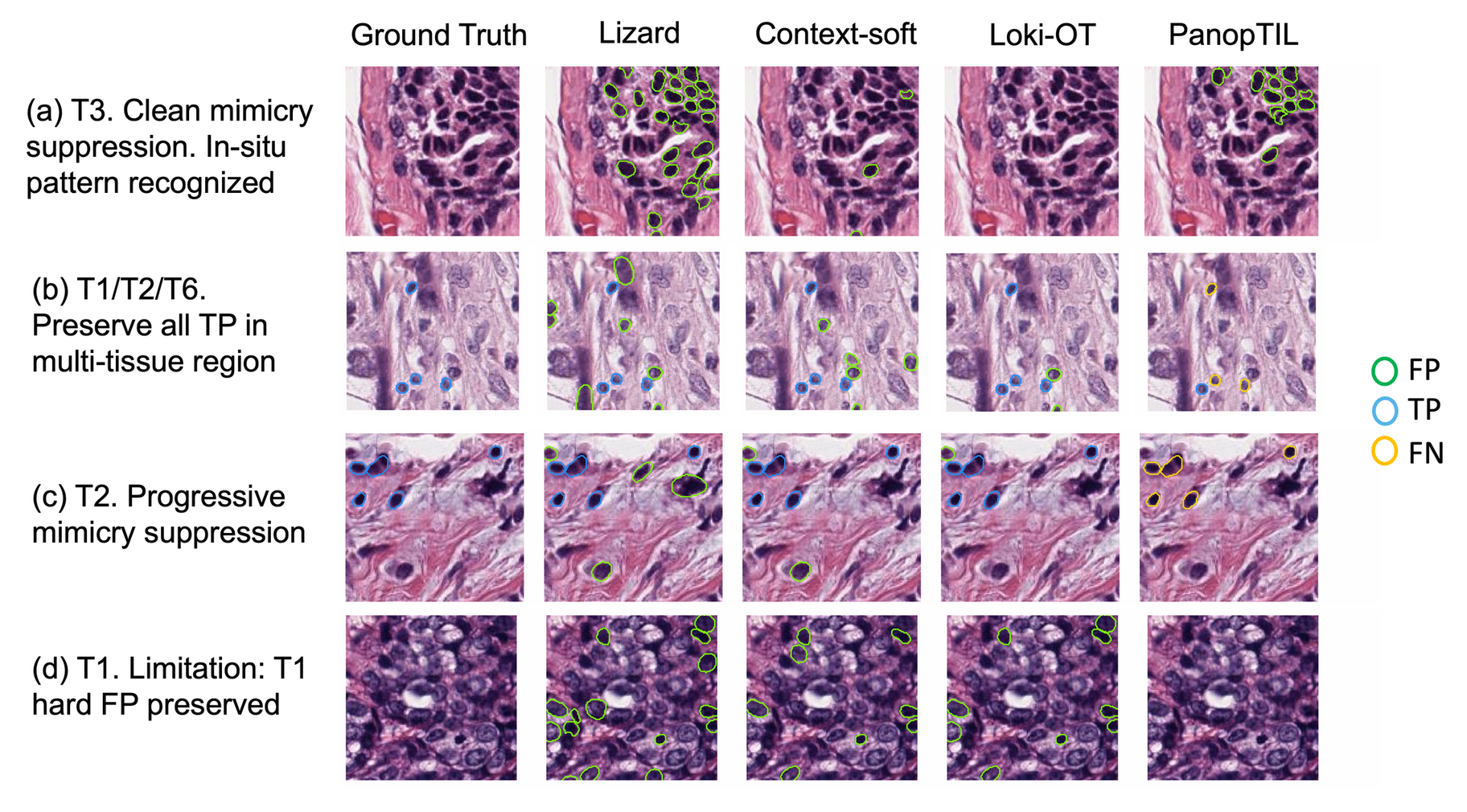}
\caption{Per-ROI cell classification on representative TCGA-BRCA test cases.
Columns: Ground Truth, Lizard, Context-soft, Loki-OT, PanopTIL;
contours: green\,=\,FP, blue\,=\,TP, yellow\,=\,FN.
(a)~T3 in-situ tumor (mimics cleanly eliminated); (b)~mixed T1/T2/T6 boundary;
(c)~T2 tumor-associated stroma; (d)~T1 invasive tumor (hard mimics persist, limitation).}
\label{fig:cases}
\end{figure}

\noindent\textbf{What does OT distillation add over Context-soft?}
Loki-OT makes a targeted trade (\textbf{Table~\ref{tab:main}}): recall drops as UOT suppresses some
true lymphocytes alongside morphological mimics, but MAE falls further---a
$32.8$\% reduction from Context-soft---and precision rises.
The transport plan forces morphologically similar cells in the same region to receive
different labels based on tissue context, moving predictions closer to the MLLM-estimated
regional density.
Qualitatively (\textbf{Fig.~\ref{fig:cases}a--c}), Loki-OT attains higher selectivity by recognizing the surrounding tissue context, suppressing morphological mimics while preserving true positives.
MAE gains are largest where mimicry is densest (per-ROI): T1 invasive tumor ($2.63$
vs.\ Context-soft $3.69$), T3 in-situ ($1.25$ vs.\ $2.05$), T4 healthy glands
($1.32$ vs.\ $1.84$).

\noindent\textbf{How does Loki-OT compare to PanopTILs?}
Using weak region-level MLLM estimates rather than dense in-domain cell-level labels,
Loki-OT outperforms PanopTILs on both F1 ($p{<}0.001$) and patient-level MAE ($p{=}0.031$; Wilcoxon, Holm-corrected; \textbf{Table~\ref{tab:main}}).
PanopTILs' lower recall may reflect its epithelium-heavy training distribution and a tendency to suppress positive predictions.
This tendency helps in epithelial regions but is detrimental in T6 inflamed stroma, where lymphocytes
are abundant: PanopTILs systematically under-predicts (per-ROI MAE $8.68$ vs.\
Loki-OT $6.70$; per-tissue-subtype MAE in App.~\ref{app:per_tissue_mae}, Table~\ref{tab:per_tissue_mae}), yielding low recall and the patient-level MAE gap.
This subtype bias is visible in case studies: PanopTILs suppresses the hard mimics that
other models predict as false positives (\textbf{Fig.~\ref{fig:cases}d}), but loses true positives in
lymphocyte-rich regions (\textbf{Fig.~\ref{fig:cases}b,c}).

\subsection{Mechanism Analysis}
\label{sec:ablation}

\textbf{Table~\ref{tab:ablation}} suggests that the two training stages address different levels of mimicry difficulty. Region-level proportional supervision (Stage~1) removes easy false positives while largely preserving true lymphocytes. Stage~2 introduces UOT-based cell reassignment, enabling harder context-dependent mimic correction through explicit region-to-cell supervision. This further reduces false positives but introduces additional TP suppression as a trade-off. OT-Teacher achieves the lowest MAE and highest FP reduction, but at severe TP loss, due to strict marginal matching with noisy MLLM priors. UOT-Teacher relaxes marginal constraints via KL penalties, trading some FP reduction for lower TP loss. Loki-OT does not simply replicate UOT-Teacher: the selective distillation threshold ($\Delta{\geq}10\%$) causes the student to fall back to Stage~1 predictions for low-deviation regions, effectively learning \emph{when} to rely on the OT teacher, yielding better selectivity and lower TP loss than UOT-Teacher.

\begin{table}[t]
\caption{Ablation on TIGER training set (135 ROIs, 80K nuclei). OT/UOT-Teacher rows are solver outputs (no distillation); Loki-OT is the deployable distilled model. \textbf{Bold} marks the training performance of the best deployable model (OT/UOT-Teacher are non-deployable).}
\label{tab:ablation}
\centering
\setlength{\tabcolsep}{3pt}
\begin{tabular}{@{}lrrrrrrr@{}}
\toprule
Method & FP & FP$\downarrow$\% & TP & TP$\downarrow$\% & Select. & F1 & MAE$\downarrow$ \\
\midrule
Baseline (Lizard) & 17,960 & --- & 6,550 & --- & --- & 0.269 & 135.6 \\
+ Context-soft & 12,046 & 32.9 & 6,275 & 4.2 & 7.84$\times$ & 0.331 & 87.5 \\
+ OT-Teacher & 6,017 & 66.5 & 4,120 & 37.1 & 1.79$\times$ & 0.328 & 48.8 \\
+ UOT-Teacher & 7,469 & 58.4 & 5,024 & 23.3 & 2.50$\times$ & 0.351 & 55.4 \\
+ Loki-OT & 7,898 & 56.0 & 5,449 & 16.8 & 3.32$\times$ & \textbf{0.366} & \textbf{56.4} \\
\bottomrule
\end{tabular}
\end{table}

\begin{figure}[!t]
\centering
\includegraphics[width=0.95\textwidth]{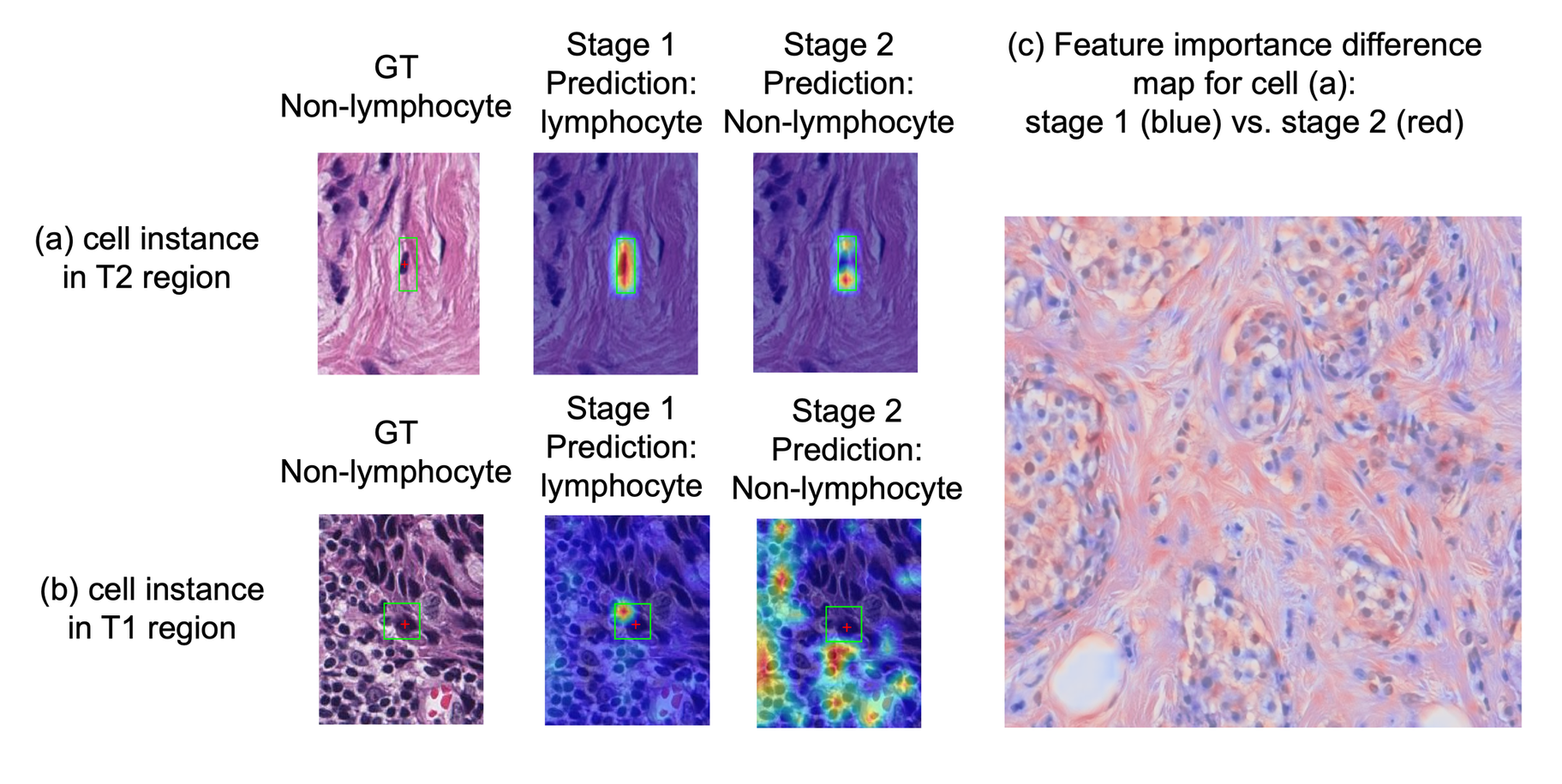}
\caption{Attention analysis at cell-token level (GradCAM) and feature level (channel attribution), for Stage~1 (Context-soft) vs.\ Stage~2 (Loki-OT).
(a,b)~Spatial GradCAM on a T2-stroma and a T1-invasive-tumor mimic.
(c)~Channel-weighted feature-importance difference map for cell~(a);
red: up-weighted by Stage~2, blue: by Stage~1.}
\label{fig:attention}
\end{figure}

\noindent\textbf{Attention analysis.} GradCAM token analysis shows that Loki-OT shifts attention from the nucleus toward the surrounding tissue, while the channel-attribution difference map indicates that Loki-OT up-weights context features (red in \textbf{Fig.~\ref{fig:attention}}).

\section{Conclusion}
\label{sec:conclusion}

We presented a cross-scale framework that transfers region-level tissue reasoning to cell-level predictions through Unbalanced Optimal Transport. On TCGA-BRCA, Loki-OT corrects lymphocyte mimicry using weak region-level MLLM estimates and achieves lower patient-level MAE than the fully supervised PanopTILs model. These findings suggest that region-level biological priors can complement dense cell annotations when errors arise from tissue context rather than detection. Our evaluation is limited to breast-cancer TIL quantification on one held-out cohort, and the asymmetric correction mechanism prioritizes precision over recall and cannot resolve errors when the baseline and prior agree. Future work will evaluate robustness to different priors and prompts and validate the framework across cancer types, cell types, and multiscale architectures.

\subsubsection*{Disclosure of Interests.} The authors have no competing interests to declare that are relevant to the content of this article.

\subsubsection*{Ethical Approval.} This study used only publicly available, de-identified human tissue datasets (TIGER challenge dataset). No new human data were collected, and institutional review board approval was therefore not required.


\newpage
\appendix

\section{MLLM Prompting Protocol}
\label{app:mllm_prompt}

Claude Sonnet 4.5 is queried once per (ROI, tissue) block in a single-turn multimodal dialog. The system and user prompts below are reproduced verbatim from the production pipeline.

\noindent\textbf{Tissue-label mapping.}
The placeholder \texttt{\{label\_description\}} in the user message is filled by matching the tissue index parsed from the (ROI, tissue) PNG filename to:
T1~\texttt{invasive\_tumor},
T2~\texttt{tumor\_stroma},
T3~\texttt{in\_situ\_tumor},
T4~\texttt{benign\_glands},
T5~\texttt{necrosis},
T6~\texttt{inflamed\_stroma},
T7~\texttt{other\_tissue}.

\noindent\textbf{System message} (sent once per query as the \texttt{system} field):

\begin{quote}\small\itshape
LYMPHOCYTE ASSESSMENT\\[4pt]
CRITICAL INSTRUCTION: ``Presence Probability'' is based primarily on ANATOMICAL CONCEPT and strict definitions (are lymphocytes within the defined compartment based on anatomical boundaries?) --- Other fields (Density, Pattern, Description) are based on VISUAL OBSERVATION combined with anatomical context (what do you see and how much?)
\end{quote}

\noindent The anatomy-vs-visual split reduces hallucination: anatomical compartment knowledge drives the binary presence call; visual evidence drives the quantitative density estimate.

\noindent\textbf{User message} (multimodal: one PNG of the ROI with the target tissue outlined in blue contour, plus the text below):

\begin{quote}\small\ttfamily
TASK: Analyze lymphocytes in the blue-outlined region\\
(labeled as \{label\_description\}):\\[4pt]
- Presence Probability (0.00-1.00):\\
\quad [likelihood based on ANATOMICAL DEFINITION]\\
- Density (if lymphocytes exist):\\
\quad [category with \% range --- VISUAL assessment]\\
\quad \textbullet{} Rare: <5\% \quad \textbullet{} Low: 5-20\%\\
\quad \textbullet{} Moderate: 20-40\% \quad \textbullet{} High: 40-70\%\\
\quad \textbullet{} Very high: >70\%\\
- Density by area (if lymphocytes exist): [\%]\\
- Density by cell count (if lymphocytes exist): [\%]\\
- Rationale: [explanation for probability assessment]\\
- Predominant location (if lymphocytes exist):\\
\quad [where most lymphocytes are]\\
- Overall pattern (if lymphocytes exist):\\
\quad [distribution --- VISUAL]\\
- Clinical interpretation: [significance]\\
- Overall confidence (0.00-1.00):\\
\quad [confidence in entire assessment]\\[4pt]
OUTPUT REQUIREMENT: Provide your response in structured JSON format.
\end{quote}

\noindent\textbf{Representative assistant response} (T6 inflamed stroma, TIGER training set):

\begin{quote}\small\ttfamily
\{\\
\quad "Presence Probability": 0.95,\\
\quad "Density": \{"category": "High", "range\_percent": "40-70\%"\},\\
\quad "Density by area": 52,\\
\quad "Density by cell count": 58,\\
\quad "Rationale": "The blue-outlined region exhibits clear lymphocytic\\
\quad\quad infiltrate with characteristic small round nuclei and high\\
\quad\quad nuclear-to-cytoplasmic ratio. Anatomically, this region is\\
\quad\quad identified as inflamed tumor stroma, where lymphocyte\\
\quad\quad infiltration is expected.",\\
\quad "Overall confidence": 0.85\\
\}
\end{quote}

\section{Per-Subtype MAE}
\label{app:per_tissue_mae}

\begin{table}[h]
\caption{Per-ROI MAE by tissue subtype (TCGA-BRCA test; lymphocytes per ROI), point estimate with 95\% bootstrap CI [low, high] ($N{=}1000$, ROI-level resampling). Due to wide, overlapping CIs in the small-$n$ subtypes (T4 $n{=}19$, T5 $n{=}58$, T3 $n{=}73$), we do not bold a single per-tissue-subtype winner. PanopTILs attains the lowest MAE in sparse-positive epithelial subtypes (T1, T3--T5, T7) by its tendency to suppress positive predictions everywhere, whereas Loki-OT is lowest in T2 and lower than PanopTILs in the lymphocyte-dense T6, consistent with its lower patient-level MAE.}
\label{tab:per_tissue_mae}
\centering
\setlength{\tabcolsep}{4pt}
\footnotesize
\begin{tabular}{@{}lcccc@{}}
\toprule
Tissue subtype ($n$) & Lizard & PanopTILs & Context-soft & Loki-OT \\
\midrule
T1 Invasive tumor (1108) & \shortstack{5.24\\{\scriptsize[4.83,5.69]}} & \shortstack{1.14\\{\scriptsize[.96,1.34]}} & \shortstack{3.69\\{\scriptsize[3.38,4.01]}} & \shortstack{2.63\\{\scriptsize[2.41,2.86]}} \\[3pt]
T2 Tumor-assoc.\ stroma (1169) & \shortstack{2.51\\{\scriptsize[2.34,2.70]}} & \shortstack{2.09\\{\scriptsize[1.91,2.28]}} & \shortstack{2.35\\{\scriptsize[2.18,2.55]}} & \shortstack{1.96\\{\scriptsize[1.80,2.12]}} \\[3pt]
T3 In-situ tumor (73) & \shortstack{3.58\\{\scriptsize[2.38,5.01]}} & \shortstack{0.67\\{\scriptsize[.25,1.22]}} & \shortstack{2.05\\{\scriptsize[1.45,2.84]}} & \shortstack{1.25\\{\scriptsize[.85,1.71]}} \\[3pt]
T4 Healthy glands (19) & \shortstack{3.47\\{\scriptsize[1.84,5.95]}} & \shortstack{0.84\\{\scriptsize[.16,1.79]}} & \shortstack{1.84\\{\scriptsize[.47,4.05]}} & \shortstack{1.32\\{\scriptsize[.32,2.84]}} \\[3pt]
T5 Necrosis (58) & \shortstack{2.84\\{\scriptsize[1.57,4.91]}} & \shortstack{0.28\\{\scriptsize[.14,.45]}} & \shortstack{3.21\\{\scriptsize[1.62,5.79]}} & \shortstack{1.74\\{\scriptsize[.98,2.91]}} \\[3pt]
T6 Inflamed stroma (624) & \shortstack{6.54\\{\scriptsize[5.97,7.15]}} & \shortstack{8.68\\{\scriptsize[7.94,9.54]}} & \shortstack{6.54\\{\scriptsize[5.96,7.16]}} & \shortstack{6.70\\{\scriptsize[6.11,7.36]}} \\[3pt]
T7 Rest (103) & \shortstack{0.51\\{\scriptsize[.35,.69]}} & \shortstack{0.29\\{\scriptsize[.18,.41]}} & \shortstack{0.41\\{\scriptsize[.26,.59]}} & \shortstack{0.31\\{\scriptsize[.19,.44]}} \\
\bottomrule
\end{tabular}
\end{table}

\end{document}